\documentclass[10pt,twocolumn,letterpaper]{article}

\usepackage{cvpr}
\usepackage{times}
\usepackage{epsfig}
\usepackage{graphicx}
\usepackage{amsmath}
\usepackage{amssymb}

\usepackage{float}
\usepackage{multirow}
\usepackage{epsfig}

\usepackage{booktabs}
\usepackage[numbers,sort&compress]{natbib}
\usepackage{color}
\usepackage[normalem]{ulem}

\usepackage[breaklinks=true,bookmarks=false]{hyperref}

\cvprfinalcopy 

\def\cvprPaperID{****} 

\begin{document}

\title{Tell Me Where to Look: Guided Attention Inference Network}

\author{
   Kunpeng Li$^{1}$, Ziyan Wu$^{3}$, Kuan-Chuan Peng$^{3}$, Jan Ernst$^{3}$ and Yun Fu$^{1,2}$\\
   $^1$Department of Electrical and Computer Engineering, Northeastern University, Boston, MA\\
   $^2$College of Computer and Information Science, Northeastern University, Boston, MA\\
   $^3$Siemens Corporate Technology, Princeton, NJ\\
   \small{\{kunpengli,yunfu\}@ece.neu.edu, \{ziyan.wu, kuanchuan.peng, jan.ernst\}@siemens.com}
}


\maketitle

\begin{abstract}



Weakly supervised learning with only coarse labels can obtain visual explanations of deep neural network such as attention maps by back-propagating gradients. These attention maps are then available as priors for tasks such as object localization and semantic segmentation. In one common framework we address three shortcomings of previous approaches in modeling such attention maps: We (1) first time make attention maps an explicit and natural component of the end-to-end training, (2) provide self-guidance directly on these maps by exploring supervision form the network itself to improve them, and (3) seamlessly bridge the gap between using weak and extra supervision if available. Despite its simplicity, experiments on the semantic segmentation task demonstrate the effectiveness of our methods. We clearly surpass the state-of-the-art on Pascal VOC 2012 val. and test set. Besides, the proposed framework provides a way not only explaining the focus of the learner but also feeding back with direct guidance towards specific tasks. Under mild assumptions our method can also be understood as a plug-in to existing weakly supervised learners to improve their generalization performance.

\end{abstract}

\section{Introduction}\label{sc:intro}

Weakly supervised learning \cite{simonyan2013deep,zeiler2014visualizing,cao2015look, zhang2016top} has recently gained much attention as a popular solution to address labeled data scarcity in computer vision. Using only image level labels for example, one can obtain attention maps for a given input with back-propagation on a Convolutional Neural Network (CNN). These maps relate to the network's response given specific patterns and tasks it was trained for. The value of each pixel on an attention map reveals to what extent the same pixel on the input image contributes to the final output of the network. It has been shown that one can extract localization and segmentation information from such attention maps without extra labeling effort.



 \begin{figure}
 \centering
 \includegraphics[width=1\columnwidth]{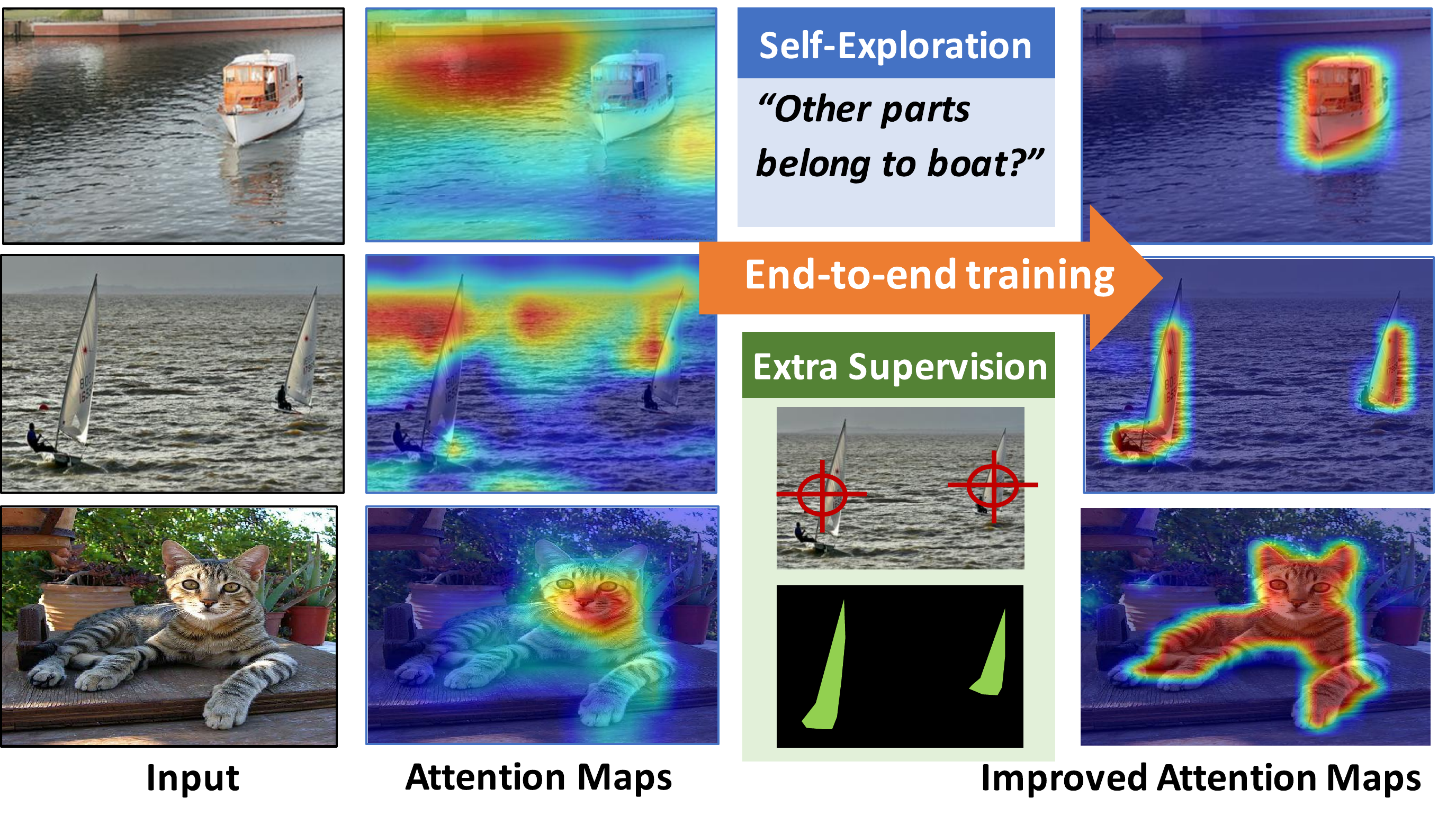} 
 \caption{The proposed Guided Attention Inference Network (GAIN) makes the network's attention on-line trainable and can plug in different kinds of supervision directly on attention maps in an end-to-end way. We explore the self-guided supervision from the network itself and propose GAIN$_{ext}$ when extra supervision are available. These guidance can optimize attention maps towards the task of interest.}
 \label{fig:teaser} 
 \end{figure} 

However, supervised by only classification loss, attention maps often only cover small and most discriminative regions of object of interest \cite{kim2017two,singh2017hide,zhou2016learning}. While these attention maps can still serve as reliable priors for tasks like segmentation \cite{kolesnikov2016seed}, having attention maps covering the target foreground objects as complete as possible can further boost the performance. To this end, several recent works either rely on  combining multiple attention maps from a network via iterative erasing steps \cite{wei2017object} or consolidating attention maps from multiple networks \cite{kim2017two}. Instead of passively exploiting trained network attention, we envision an end-to-end framework with which task-specific supervision can be directly applied on attention maps during training stage.


On the other hand, as an effective way to explain the network's decision, attention maps can help to find restrictions of the training network. For instance in an object categorization task with only image-level object class labels, we may encounter a pathological bias in the training data when the foreground object incidentally always correlates with the same background object (also pointed out in \cite{grad-cam}). Figure \ref{fig:teaser} shows the example class "boat" where there may be bias towards water as a distractor with high correlation. In this case the training has no incentive to focus attention only on the foreground class and generalization performance may suffer when the testing data does not have the same correlation ("boats out of water"). While there have been attempts to remove this bias by re-balancing the training data, we instead propose to model the attention map \emph{explicitly} as part of the training. As one benefit of this we are able to control the attention explicitly and can put manual effort in providing minimal supervision of attention rather than re-balancing the data set. While it may not always be clear how to manually balance data sets to avoid bias, it is usually straightforward to guide attention to the regions of interest. We also observe that our explicit self-guided attention model already improves the generalization performance even without extra supervision.

Our contributions are: (a) A method of using supervision directly on attention maps during training time while learning a weakly labeled task; (b) A scheme for self-guidance during training that forces the network to focus attention on the object holistically rather than only the most discriminative parts; (c) Integration of direct supervision and self-guidance to seamlessly scale from using only weak labels to using full supervision in one common framework.

Experiments using semantic segmentation as task of interest show that our approach achieves mIoU 55.3\% and 56.8\%, respectively on the \emph{val} and \emph{test} of the PASCAL VOC 2012 segmentation benchmark. It also confidently surpasses the comparable state-of-the-art when limited pixel-level supervision is used in training with an mIoU of 60.5\% and 62.1\% respectively. To the best of our knowledge these are the new state-of-the-art results under weak supervision.

%
%
%
%
%

\section{Related work}\label{sc:related_work}

Since deep neural networks have achieved great success in many areas \cite{gong2017learning,zhang2017image}, various methods have been proposed to try to explain this black box \cite{simonyan2013deep,zeiler2014visualizing,cao2015look,zhou2014object,zhang2017mdnet}. Visual attention is one way that tries to explain which region of the image is responsible for network's decision. In \cite{simonyan2013deep,zeiler2014visualizing,springenberg2015striving}, error backpropagation based methods are used for visualizing relevant regions for a predicted class or the activation of a hidden neuron. In \cite{cao2015look}, a feedback CNN architecture is proposed for capturing the top-down attention mechanism that can successfully identify task relevant regions. CAM \cite{zhou2016learning} shows that replacing fully-connected layers with an average pooling layer can help generate coarse class activation maps that highlight task relevant regions. Inspired by a top-down human visual attention model, \cite{zhang2016top} proposes a new backpropagation scheme, called Excitation Backprop, to pass along top-down signals downwards in the network hierarchy. Recently, Grad-CAM \cite{grad-cam} extends the CAM to various off-the-shelf available architectures for tasks including image classification, image captioning and VQA providing faithful visual explanations for possible model decisions. Different from all these methods that are trying to explain the network, we first time build up an end-to-end model to provide supervision directly on these explanations, specifically network's attention here. We validate these supervision can guide the network focus on the regions we expect and benefit the corresponding visual tasks.

\begin{figure*}
\centering
\includegraphics[width=0.95\linewidth]{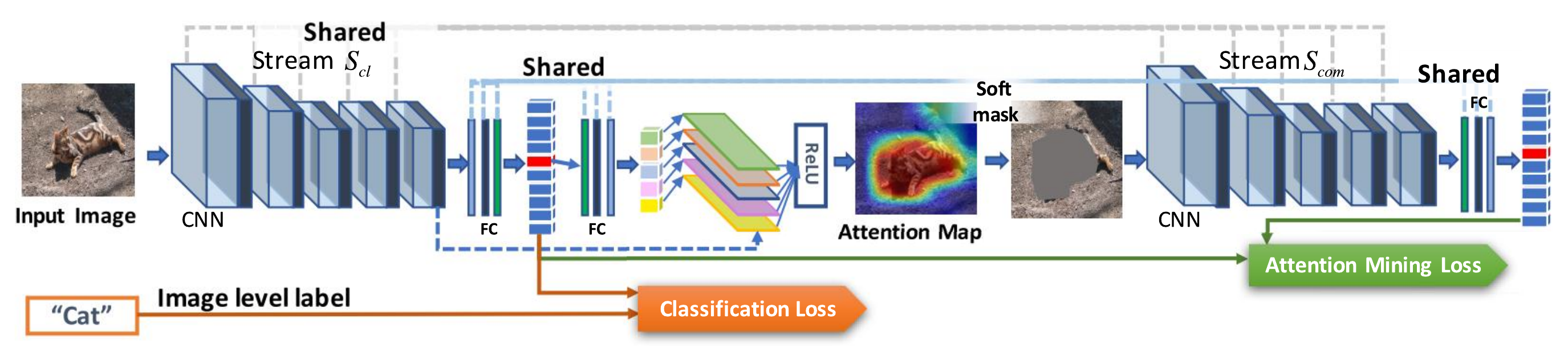} 
\caption{GAIN has two streams of networks, ${S_{cl}}$ and ${S_{am}}$, sharing parameters. ${S_{cl}}$ aims to find out regions that help to recognize the object and ${S_{am}}$ tries to make sure all these regions contributing to this recognition have been discovered. The attention map is on-line generated and trainable by the two loss functions jointly.}
\label{fig:self_supervised} 
\end{figure*}

Many methods heavlily rely on the location information provided by the network's attention. Learning from only the image-level labels, attention maps of a trained classification network can be used for weakly-supervised object localization \cite{zhou2016learning,oquab2015object}, anomaly localization, scene segmentation \cite{kolesnikov2016seed} and etc. However, only trained with classification loss, the attention map only covers small and most discriminative regions of the object of interest, which deviates from the requirement of these tasks that needs to localize dense, interior and complete regions. To mitigate this gap, \cite{singh2017hide} proposes to hide patches in a training image randomly, forcing the network to seek other relevant parts when the most discriminative part is hidden. This approach can be considered as a way to augment the training data, and it has strong assumption on the size of foreground objects (i.e., the object size vs. the size of the patches). In \cite{wei2017object}, use the attention map of a trained network to erase the moset discriminative regions of the original input image. And the repeat this erase and discover action to the erased image for several steps and combine attention maps of each step to get a more complete attention map. Similarly, \cite{kim2017two} uses a two-phase learning stratge and combine attention maps of the two networks to get a more complete region for the object of interest. In the first step, a conventional fully convolutional network (FCN) \cite{long2015fully} is trained to find the most discriminative parts of an image. Then these most salient parts are used to supresse the feature map of the secound network to force it to focus on the next most important parts. However, these methods either rely on combinations of attention maps of one trained network for different erased steps or attentions of different networks. The single network's attention still only locates on the most discriminative region. Our proposed GAIN model is fundamentally different from the previous approaches. Since our models can provide supervision directly on network's attention in an end-to-end way, which can not be done by all the other methods \cite{grad-cam,zhou2016learning,wei2017object,zhang2016top,singh2017hide,kim2017two}, we design different kinds of loss functions to guide the network focus on the whole object of interest. Therefore, we do not need to do several times erasing or combine attention maps. The attention of our single trained network is already more complete and improved.

Identifying bias in datasets \cite{torralba2011unbiased} is another important usage of the network attention. \cite{grad-cam} analyses the location of attention maps of a trained model to find out the dataset bias, which helps them to build a better unbiased dataset. However, in practical applications, it is hard remove all the bias of the dataset and time-consuming to build a new dataset. How to garantee the generalization ability of the learned network is still challenging. Different from the existing methods, our model can fundamentally solve this problem by providing supervision directly on network's attention and guiding the network to focus on the areas critical to the task of interest, therefore is robust to dataset bias.

\section{Proposed method --- GAIN}

Since attention maps reflect the areas on input image which support the network's prediction, we propose the guided attention inference networks (GAIN), which aims at supervising attention maps when we train the network for the task of interest. In this way, the network's prediction is based on the areas which we expect the network to focus on. We achieve this by making the network's attention trainable in an end-to-end fashion, which hasn't been considered by any other existing works \cite{grad-cam,zhou2016learning,wei2017object,zhang2016top,singh2017hide,kim2017two}. In this section, we describe the design of GAIN and its extensions tailored towards tasks of interest.


\subsection{Self-guidance on the network attention} \label{section:method_self}

As mentioned in Section \ref{sc:related_work}, attention maps of a trained classification network can be used as priors for weakly-supervised semantic segmentation methods. However, purely supervised by the classification loss, attention maps usually only cover small and most discriminative regions of object of interest. These attention maps can serve as reliable priors for segmentation but a more complete attention map can certainly help improving the overall performance.

To solve this issue, our GAIN builds constrains directly on the attention map in a regularized bootstrapping fashion. As shown in Figure \ref{fig:self_supervised}, GAIN has two streams of networks, classification stream ${S_{cl}}$ and attention mining ${S_{am}}$, which share parameters with each other. The constrain from stream ${S_{cl}}$ aims to find out regions that help to recognize classes. And the stream ${S_{am}}$ is making sure that all regions which can contribute to the classification decision will be included in the network's attention. In this way, attention maps become more complete, accurate and tailored for the segmentation task. The key here is that we make the attention map can be on-line generated and trainable by the two loss functions jointly.

Based on the fundemantal framework of Grad-CAM \cite{grad-cam}, we streamlined the generation of attention map.  An attention map corresponding to the input sample can be obtained within each inference so it becomes trainable in training statge. In stream ${S_{cl}}$, for a given image $I$, let $f_{l,k}$ be the activation of unit $k$ in the $l$-th layer. For each class $c$ from the ground-truth label, we compute the gradient of the score $s^c$ corresponding to class $c$, with respect to activation maps of $f_{l,k}$. These gradients flowing back will pass through a global average pooling layer \cite{lin2013network} to obtain the neuron importance weights $w_{l,k}^c$ as defined in Eq. \ref{eq:improtance_weights}.

\begin{equation}
\label{eq:improtance_weights}
w_{l,k}^c = {\mathop{\rm GAP}\nolimits}\left( {\frac{{\partial {s^c}}}{{\partial {f_{l,k}}}}} \right),
\end{equation}
where ${\mathop{\rm GAP}\nolimits} \left(  \cdot  \right)$ means global average pooling operation.

Here, we do not update parameters of the network after obtaining the $w_{l,k}^c$ by back-propagation. Since $w_{l,k}^c$ represents the importance of activation map $f_{l,k}$ supporting the prediction of class $c$, we then use weights matrix $w^c$ as the kernel and apply 2D convolution over activation maps matrix $f_{l}$ in order to integrate all activation maps, followed by a ReLU operation to get the attention map $A^c$ with Eq. \ref{eq:attention_map}. The attention map is now on-line trainable and constrains on $A^c$ will influence the network's learning:

\begin{equation}
\label{eq:attention_map}
{A^c} = {\mathop{\rm ReLU}\nolimits} \left( {{\mathop{\rm conv}\nolimits} \left( {{f_l},{w^c}} \right)} \right),
\end{equation}
where $l$ is the representation from the last convolutional layer whose features have the best compromise between high-level semantics and detailed spatial information \cite{simonyan2013deep}. The attention map has the same size as the convolutional feature maps ($14 \times 14$ in the case of VGG \cite{simonyan2014very}).

\begin{figure*}
\centering
\includegraphics[width=0.98\linewidth]{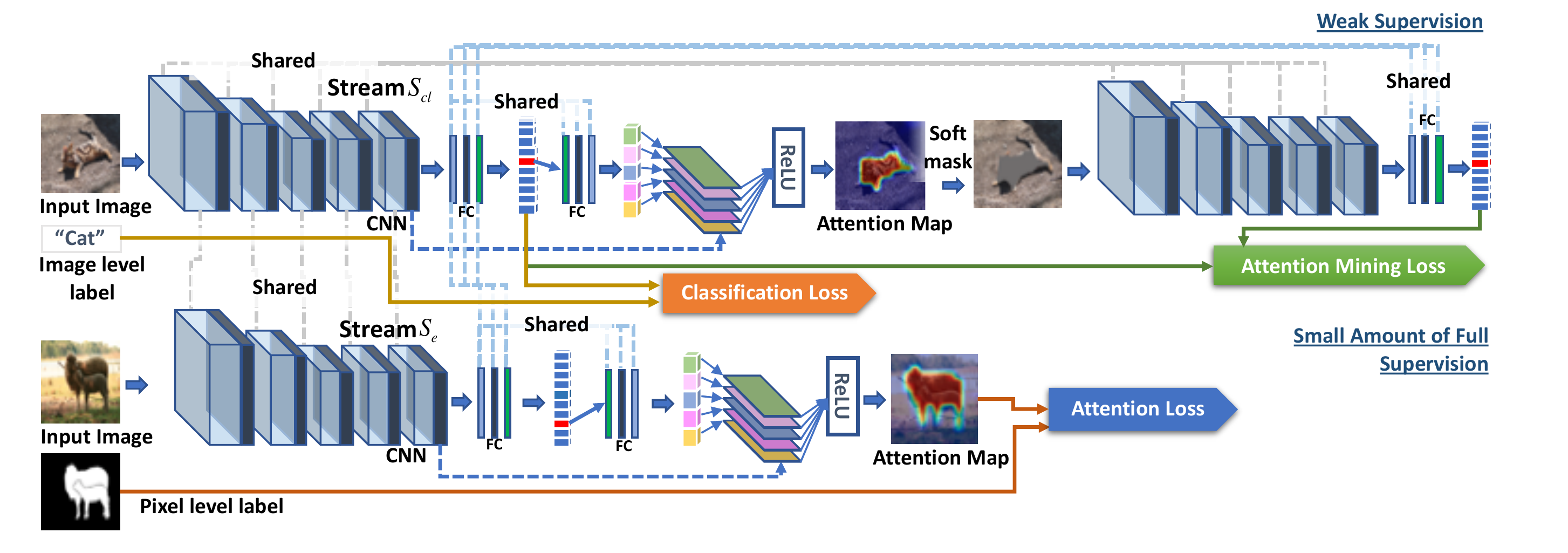} 
\caption{Framework of the GAIN$_{ext}$. Pixel-level annotations are seamlessly integrated into the GAIN framework to provide direct supervision on attention maps optimizing towards the task of semantic segmentation.}
\label{fig:human_supervised} 
\end{figure*} 

We then use the trainable attention map $A^c$ to generate a soft mask to be applied on the original input image, obtaining $I^ {*c}$ using Eq. \ref{eq:croped_img}. $I^ {*c}$ represents the regions beyond the network's current attention for class $c$.

\begin{equation}
\label{eq:croped_img}
{I^ {*c} } = I - \left( T\left( {{A^c}} \right) \odot I \right),
\end{equation}
where $\odot$ denotes element-wise multiplication. $T\left( {{A^c}} \right)$ is a masking function based on a thresholding operation. In order to make it derivable, we use Sigmoid function as an approximation defined in Eq. \ref{eq:crop_function}.

\begin{equation}
\label{eq:crop_function}
T\left( {{A^c}} \right) = \frac{1}{{1 + \exp \left( { - \omega \left( {{A^c} -  \boldsymbol{\sigma} } \right)} \right)}}，
\end{equation}
where $\boldsymbol{\sigma}$ is the threshold matrix whose elements all equal to $\sigma$. $\omega$ is the scale parameter ensuring $T\left( {{A^c}} \right){}_{i,j}$ approximately equals to 1 when ${A^c}_{i,j}$ is larger than $\sigma$, or to 0 otherwise.

$I^ {*c}$ is then used as input of stream ${S_{am}}$ to obtain the class prediction score. Since our goal is to guide the network to focus on all parts of the class of interest, we are enforcing $I^ {*c}$ to contain as little feature belonging to the target class as possible, i.e. regions beyond the high-responding area on attention map area should include ideally not a single pixel that can trigger the network to recognize the object of class $c$. From the loss function perspective it is trying to minimize the prediction score of $I^ {*c}$ for class $c$. To achieve this, we design the loss function called Attention Mining Loss as in Eq. \ref{eq:loss_croped}.

\begin{equation}
\label{eq:loss_croped}
{L_{am}} = \frac{1}{n}\sum\limits_c {{s^c}({I^{*c}})},
\end{equation}
where ${{s^c}({I^{*c}})}$ denotes the prediction score of $I^ {*c}$ for class $c$.  $n$ is the number of ground-truth class labels for this image $I$. 

As defined in Eq. \ref{eq:self_guidance_loss}, our final self-guidance loss $L_{self}$ is the summation of the classification loss $L_{cl}$ and $L_{am}$.

\begin{equation}
\label{eq:self_guidance_loss}
{L_{self}} = {L_{cl}} + \alpha{L_{am}},
\end{equation}
where ${L_{cl}}$ is for multi-label and multi-class classification and we use a multi-label soft margin loss here. Alternative loss functions can be use for specific tasks. $\alpha$ is the weighting parameter. We use $\alpha = 1$ in all of our experiments. 

With the guidance of $L_{self}$, the network learn to extend the focus area on input image contributing to the recognition of target class as much as possible, such that attention maps are tailored towards the task of interest, i.e. semantic segmentation. We demonstrate the efficacy of GAIN with self guidance in Sec.~\ref{sc:seg_exp}.

\subsection{GAIN$_{ext}$: integrating extra supervision} \label{section:method_human_guided}

In addition to letting networks explore the guidance of the attention map by itself, we can also tell networks which part in the image they should focus on by using a small amount of extra supervision to control the attention map learning process, so that to be tailored for the task of interest. Based on this idea of imposing additional supervision on attention maps, we introduce the extension of GAIN: GAIN$_{ext}$, which can seamlessly integrate extra supervision in our weakly supervised learning framework. We demonstrate using the self-guided GAIN framework improving the weakly-supervised semantic segmentation task as shown in Sec.~\ref{sc:seg_exp}. Furthermore, we can also apply GAIN$_{ext}$ to guide the network to learn features robust to dataset bias and improve its generalizability when the testing data and training data are drawn from very different distributions.

Following Sec.~\ref{section:method_self}, we still use the weakly supervised semantic segmentation task as an example application to explain the GAIN$_{ext}$. The way to generate trainable attention maps in GAIN$_{ext}$ during training stage is the same as that in the self-guided GAIN. In addition to $L_{cl}$ and $L_{am}$, we design another loss $L_e$ based on the given external supervision. We define $L_e$ as:

\begin{equation}
\label{eq:human_guided_loss}
{L_e} = \frac{1}{n}\sum\limits_c {{{\left( {{A^c} - {H^c}} \right)}^2}},
\end{equation}
where ${H^c}$ denotes the extra supervision, e.g. pixel-level segmentation masks in our example case.

Since generating pixel-level segmentation maps is extremely time consuming, we are more interested in finding out the benefits of using only a very small amount of data with external supervision, which fits perfectly with the GAIN$_{ext}$ framework shown in Figure~\ref{fig:human_supervised}, where we add an external stream ${S_e}$, and these three streams share all parameters. Input images of stream ${S_e}$ include both image-level labels and pixel-level segmentation masks. One can use only very small amount of pixel-level labels through stream ${S_e}$ to already gain performance improvement with GAIN$_{ext}$ (in our experiments with GAIN$_{ext}$, only 1$\sim$10\% of the total labels used in training are pixel-level labels). The input of the stream $S_{cl}$ includes all the images in the training set with only image-level labels.

The final loss function, $L_{ext}$, of GAIN$_{ext}$ is defined as follows:
\begin{equation}
\label{eq:human_guidance_loss_final}
L_{ext} = L_{cl} + \alpha L_{am} + \omega L_e,
\end{equation} 
where $L_{cl}$ and $L_{am}$ are defined in Sec.~\ref{section:method_self}, and $\omega$ is the weighting parameter depending on how much emphasis we want to place on the extra supervision (we use $\omega  = 10$ in our experiments).

GAIN$_{ext}$ can also be easily modified to fit other tasks. Once we get activation maps $f_{l,k}$ corresponding to the network's final output, we can use $L_e$ to guide the network to focus on areas critical to the task of interest. In Sec.~\ref{section:human_guided_classification_experiment}, we show an example of such modification to guide the network to learn features robust to dataset bias and improve its generalizability. In that case, extra supervision is in the form of bounding boxes. 

\setlength\tabcolsep{2pt}
 \begin{table}
 \small
 \begin{center}
 \begin{tabular}{lccc}
 \hline
 Methods & Training Set & \emph{val.}  &  \emph{test}\\
  & & (mIoU) & (mIoU) \\
 \hline\hline
 \multicolumn{3}{l}{Supervision: Purely Image-level Labels } \\
 CCNN \cite{pathak2015constrained} & 10K weak & 35.3 & 35.6\\
 MIL-sppxl \cite{pinheiro2015image} & 700K weak & 35.8 & 36.6\\
 EM-Adapt \cite{papandreou2015weakly} & 10K weak & 38.2 & 39.6\\
 DCSM \cite{shimoda2016distinct} & 10K weak & 44.1 & 45.1\\
 BFBP \cite{saleh2016built} & 10K weak & 46.6 & 48.0\\
 STC \cite{wei2017stc} & 50K weak & 49.8 & 51.2\\
 AF-SS \cite{qi2016augmented} & 10K weak & 52.6 & 52.7\\
 CBTS-cues \cite{roy2017combining} & 10K weak & 52.8 & 53.7\\
 TPL \cite{kim2017two} & 10K weak & 53.1 & 53.8\\
 AE-PSL \cite{wei2017object} & 10K weak & 55.0 & 55.7\\
 SEC \cite{kolesnikov2016seed} (baseline) & 10K weak & 50.7 & 51.7\\
 GAIN (ours) & 10K weak & 55.3 & 56.8\\
 \hline
 \multicolumn{3}{l}{Supervision: Image-level Labels } \\
 \multicolumn{3}{l}{(* Implicitly use pixel-level supervision)} \\
 MIL-seg* \cite{pinheiro2015image} & 700K weak + 1464 pixel & 40.6 & 42.0\\
 TransferNet* \cite{hong2016learning} & 27K weak + 17K pixel & 51.2 & 52.1\\
 AF-MCG* \cite{qi2016augmented} & 10K weak + 1464 pixel & 54.3 & 55.5\\
 GAIN$_{ext}$* (ours) & 10K weak + 200 pixel & 58.3 & 59.6\\
 GAIN$_{ext}$* (ours) & 10K weak + 1464 pixel & 60.5 & 62.1\\
 \hline
 \end{tabular}
 \end{center}
 \caption{Comparison of weakly supervised semantic segmentation methods on VOC 2012 \emph{segmentation val.} set and \emph{segmentation test} set. \textbf{weak} denotes image-level labels and \textbf{pixel} denotes pixel-level labels. \textit{Implicitly use pixel-level supervision} is a protocol we followed as defined in \cite{wei2017object}, that pixel-level labels are only used in training priors, and only weak labels are used in the training of segmentation framework, e.g. SEC \cite{kolesnikov2016seed} in our case.}
 \label{table:voc_val_test}
 \end{table}
 
\section{Semantic segmentation experiments}\label{sc:seg_exp}

To verify the efficacy of GAIN, following Sec.~\ref{section:method_self} and \ref{section:method_human_guided}, we use the weakly supervised semantic segmentation task as the example application. The goal of this task is to classify each pixel into different categories. In the weakly supervised setting, most of recent methods~\cite{kolesnikov2016seed,kim2017two,wei2017object} mainly rely on localization cues generated by models trained with only image-level labels and consider other constraints such as object boundaries to train a segmentation network. Therefore, the quality of localization cues is the key of these methods' performance.

Compared with attention maps generated by the state-of-the-art methods~\cite{grad-cam,long2015fully,zhou2016learning} which only locate the most discriminative areas, GAIN guides the network to focus on entire areas representing the class of interest, which can improve the performance of weakly supervised segmentation. To verify this, we adopt our attention maps to SEC~\cite{kolesnikov2016seed}, which is one of the state-of-the-art weakly supervised semantic segmentation methods. SEC defines three key constrains: \emph{seed}, \emph{expand} and \emph{constrain}, where \emph{seed} is a module to provide localization cues $C$ to the main segmentation network $N$ such that the segmentation result of $N$ is supervised to match $C$. Note that SEC is not a dependency of GAIN. It is used here in order to evaluate improvements brought by attention priors produced by GAIN. In principal it can be replaced by other segmentation frameworks for this application. Following SEC~\cite{kolesnikov2016seed}, our localization cues are obtained by applying a thresholding operation to attention maps generated by GAIN: for each per-class attention map, all pixels with a score larger than 20\% of the maximum score are selected. We use \cite{liu2016dhsnet} to get background cues and then train the SEC model to generate segmentation results using the same inference procedure, as well as parameters of CRF\cite{CRFkrahenbuhl2011efficient}.
 
\subsection{Dataset and experimental settings}\label{sc:Experiment_setting}

\textbf{Dataset and evaluation metrics.} We evaluate our results on the PASCAL VOC 2012 image segmentation benchmark~\cite{everingham2010pascal}, which has 21 semantic classes, including the background. The images are split into three sets: training, validation, and testing (denoted as train, val, and test) with 1464, 1449, and 1456 images, respectively. Following the common setting~\cite{chen14semantic,kolesnikov2016seed}, we use the augmented training set provided by~\cite{hariharan2011semantic}. The resulting training set has 10582 weakly annotated images which we use to train our models. We compare our approach with other approaches on both the val and test sets. The ground truth segmentation masks for the test set are not publicly available, so we use the official PASCAL VOC evaluation server to obtain the quantitative results. For the evaluation metric, we use the standard one for the PASCAL VOC 2012 segmentation --- mean intersection-over-union (mIoU).

\textbf{Implementation details.} We use the VGG~\cite{simonyan2014very} pretrained from the ImageNet~\cite{deng2009imagenet} as the basic network for GAIN to generate attention maps. We use Pytorch~\cite{Pytorch} to implement our models. We set the batch size to 1 and learning rate to ${10^{ - 5}}$. We use the stochastic gradient descent (SGD) to train the networks and terminate after 35 epochs. For the weakly-supervised segmentation framework, following the setting of SEC~\cite{kolesnikov2016seed}, we use the DeepLab-CRFLargeFOV~\cite{chen14semantic}, which is a slightly modified version of the VGG network~\cite{simonyan2014very}. Implemented using Caffe~\cite{jia2014caffe}, DeepLab-CRFLargeFOV~\cite{chen14semantic} takes the inputs of size 321$\times$321 and produces the segmentation masks of size 41$\times$41. Our training procedure is the same as~\cite{kolesnikov2016seed} at this stage. We run the SGD for 8000 iterations with the batch size of 15. The initial learning rate is ${10^{ - 3}}$ and it decreases by a factor of 10 for every 2000 iterations. 


\subsection{Comparison with state-of-the-art} \label{sc:Experiment_comparisons}

We compare our methods with other state-of-the-art weakly supervised semantic segmentation methods with image-level labels. Following~\cite{wei2017object}, we separate them into two categories. For methods purely using image-level labels, we compare our GAIN-based SEC (denoted as GAIN in the table) with SEC \cite{kolesnikov2016seed}, AE-PSL \cite{wei2017object}, TPL \cite{kim2017two}, STC \cite{wei2017stc} and etc. For another group of methods, implicitly using pixel-level supervision means that though these methods train the segmentation networks only with image-level labels, they use some extra technologies that are trained using pixel-level supervision. Our GAIN$_{ext}$-based SEC (denoted as GAIN$_{ext}$ in the table) lies in this setting because it uses a very small amount of pixel-level labels to further improve the network's attention maps and doesn't rely on any pixel-level labels when training the SEC segmentation network. Other methods in this setting like AF-MCG \cite{zhou2016learning}, TransferNet \cite{hong2016learning} and MIL-seg \cite{pinheiro2015image} are included for comparison. Table \ref{table:voc_val_test} shows results on PASCAL VOC 2012 \emph{segmentation val.} set and \emph{segmentation test.} set.

\begin{figure}
\centering
\includegraphics[width=3.2 in]{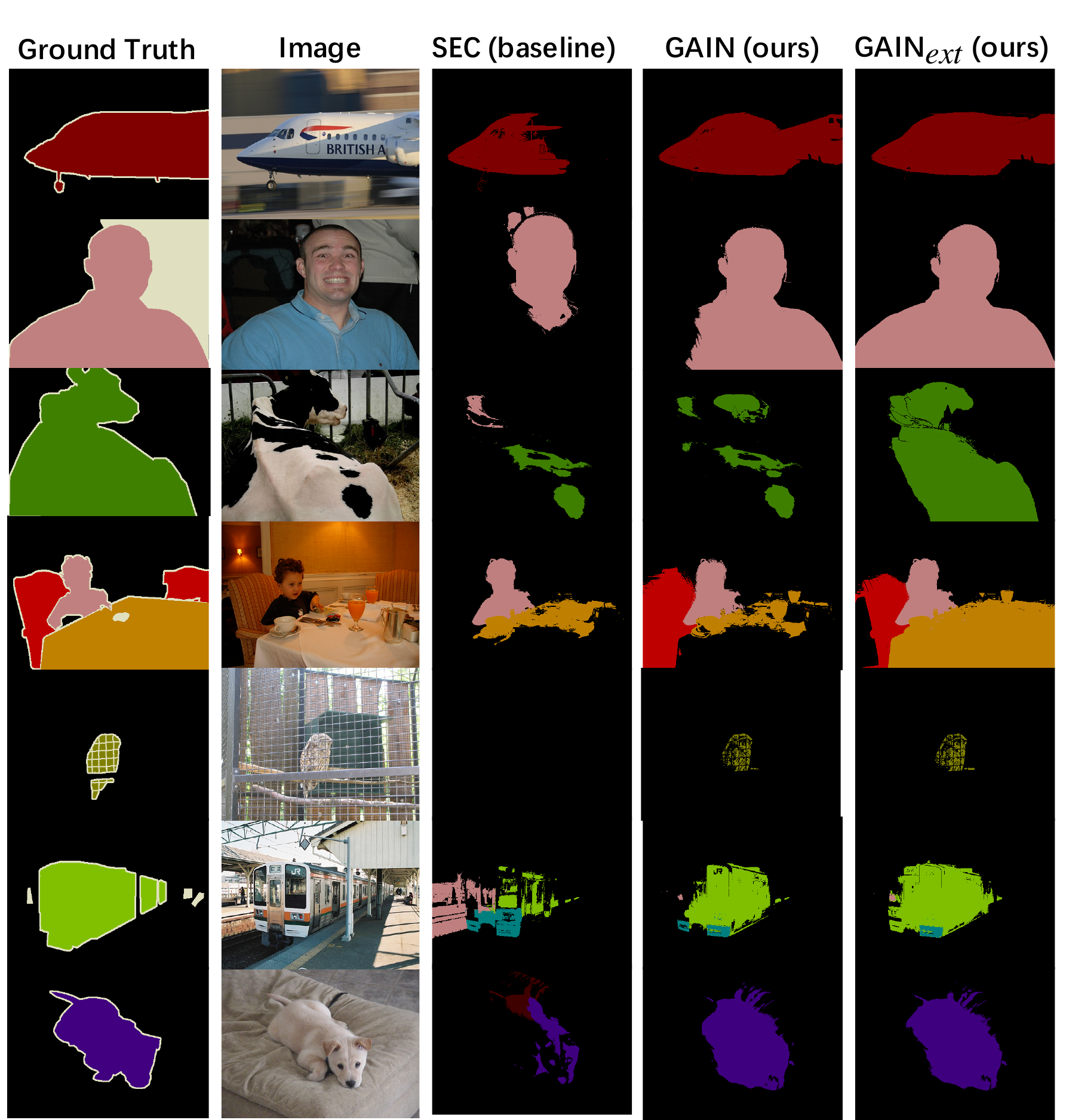} 
\caption{Qualitive results on Pascal VOC 2012 \emph{segmentation val.} set. They are generated by SEC (our baseline framework), our GAIN-based SEC and GAIN$_{ext}$-based SEC implicitly using 200 randomly selected (2\%) extra supervision.}
\label{fig:Q_seg} 
\end{figure}

\begin{figure*}
\centering
\includegraphics[width=0.95\linewidth]{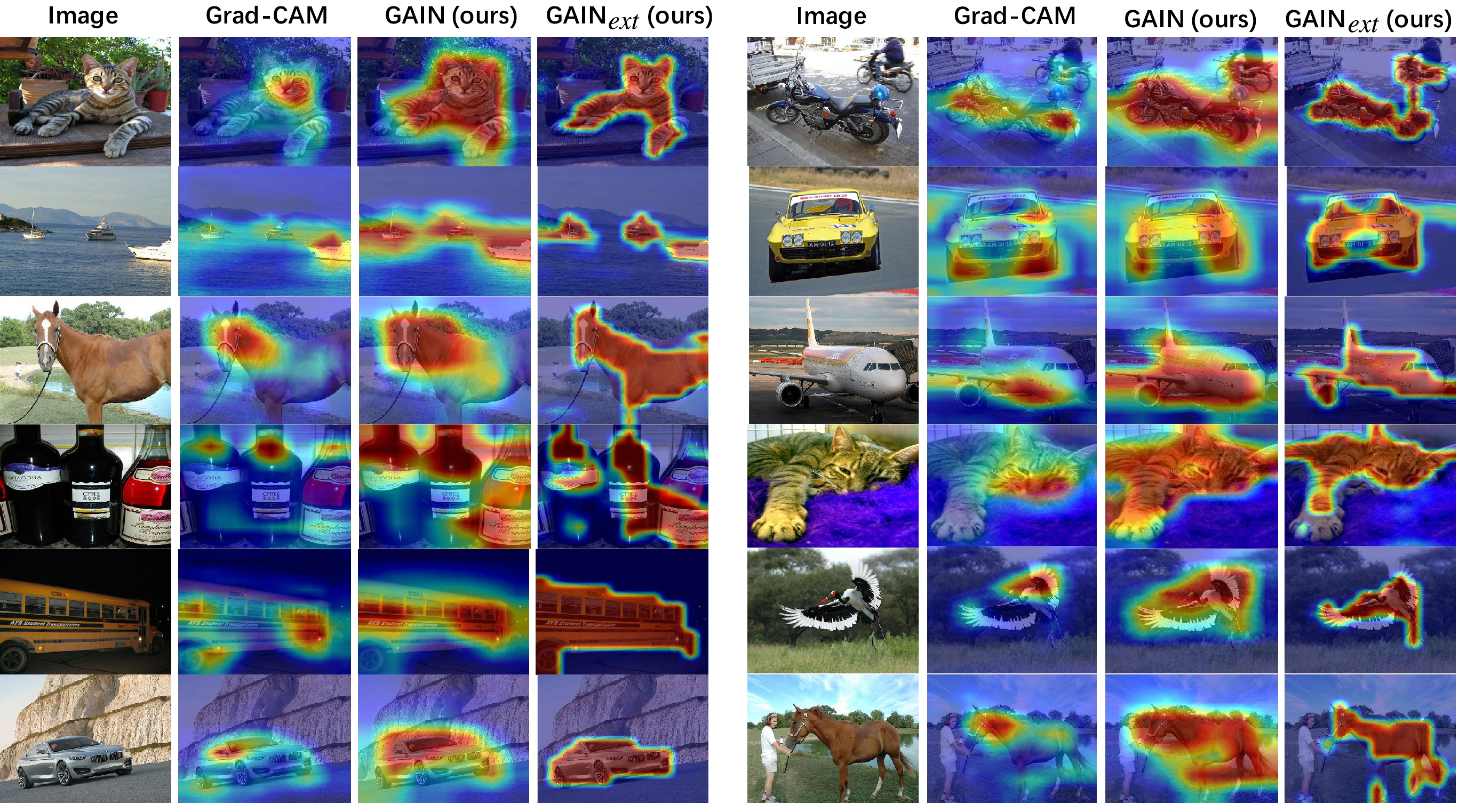} 
\caption{Qualitative results of attention maps generated by Grad-CAM \cite{grad-cam}, our GAIN and GAIN$_{ext}$ using 200 randomly selected (2\%) extra supervision.}
\label{fig:Q_heatmaps} 
\end{figure*} 
Among the methods purely using image-level labels, our GAIN-based SEC achieves the best performance with 55.3\% and 56.8\% in mIoU on these two sets, outperforming the SEC~\cite{kolesnikov2016seed} baseline by 4.6\% and 5.1\%. Furthermore, GAIN outperforms AE-PSL~\cite{wei2017object} by 0.3\% and 1.1\%, and outperforms TPL~\cite{kim2017two} by 2.2\% and 3.0\%. These two methods are also proposed to cover more areas of the class of interest in attention maps. However, they either rely on the combinations of attention maps of one trained network for different erasing steps~\cite{wei2017object} or attention maps from different networks~\cite{kim2017two}. Compared with them, our GAIN makes the attention map trainable and uses $L_{self}$ loss to guide attention maps to cover entire class of interest. The design of GAIN already makes the attention map of a single network cover more areas belonging to the class of interest without the need to do iterative erasing or combining attention maps from different networks, as proposed in~\cite{kim2017two,wei2017object}.


By implicitly using pixel-level supervision, our GAIN$_{ext}$-based SEC achieves 58.3\% and 59.6\% in mIoU when we use 200 randomly selected images with pixel-level labels (2\% data of the whole dataset) as the pixel-level supervision. It already performs 4\% and 4.1\% better than AF-MCG~\cite{zhou2016learning}, which relies on the MCG generator~\cite{arbelaez2014multiscale} trained in a fully-supervised way on the PASCAL VOC. When the pixel-level supervision increases to 1464 images for our GAIN$_{ext}$, the performance jumps to 60.5\% and 62.1\%, which is a new state-of-the-art for this challenging task on a competitive benchmark. Figure~\ref{fig:Q_seg} shows some qualitative example results of semantic segmentation, indicating that GAIN-based methods help to discover more complete and accurate areas of classes of interest based on the improvement of attention maps. Specifically, GAIN-based methods discover either other parts of objects of interest or new instances which can not be found by the baseline.

\begin{table}
 \centering
 \begin{tabular}{lccc}
 \hline
 Methods & Training Set & \emph{val.}  &  \emph{test}\\
 \hline
 SEC[11] w/o. CRF & 10K weak & 44.79 & 45.43\\
 GAIN w/o. CRF & 10K weak  & 50.78 & 51.76\\
 GAIN$_{ext}$ w/o. CRF & 10K weak + 1464 pixel & 54.77 & 55.72\\
 \hline
 \end{tabular}
 \caption{Semantic segmentation results without CRF on VOC 2012 \textit{segmentation val.} and \textit{test} sets. Numbers shown are mIoU.}
 \label{table:voc_ablation_results}
 \end{table}

We also show qualitative results of attention maps generated by GAIN-base methods in Figure~\ref{fig:Q_heatmaps}, where GAIN covers more areas belonging to the class of interest compared with the Grad-CAM~\cite{grad-cam}. With only 2\% of the pixel-level labels, the GAIN$_{ext}$ covers more complete and accurate areas of the class of interest as well as less background areas around the class of interest (for example, the sea around the ships and the road under the car in the second row of Figure~\ref{fig:Q_heatmaps}).

\textbf{More discussion of the GAIN$_{ext}$} We are interested in finding out the influence of different amount of pixel-level labels on the performance. Following the same setting in Sec.~\ref{sc:Experiment_setting}, we add more randomly selected pixel-level labels to further improve attention maps and adopt them in the SEC~\cite{kolesnikov2016seed}. From the results in Table~\ref{table:voc_val_different_percent_supervision}, we find that the performance of the GAIN$_{ext}$ improves when more pixel-level labels are provided to train the network generating attention maps. Again, there are no pixel-level labels used to train the SEC segmentation framework. 

We also evaluate performance on VOC 2012 \textit{seg. val.} and \textit{seg. test} datasets without CRF as shown in Table~\ref{table:voc_ablation_results}.

\begin{table}
\begin{center}
\begin{tabular}{cc}
\hline
Training Set & mIoU \\
\hline\hline
 10K weak + 200 pixel & 58.3\\
 10K weak + 400 pixel & 59.4\\
 10K weak + 900 pixel & 60.2\\
 10K weak + 1464 pixel & 60.5\\
\hline
\end{tabular}
\end{center}
\caption{Results on Pascal VOC 2012 \emph{segmentation val}. set with our GAIN$_{ext}$-based SEC implicitly using different amount of pixel-level supervision for the attention map learning process.}
\label{table:voc_val_different_percent_supervision}
\end{table}

\section{Guided learning with biased data}\label{section:human_guided_classification_experiment}

\begin{figure*}
\centering
\includegraphics[width=1\linewidth]{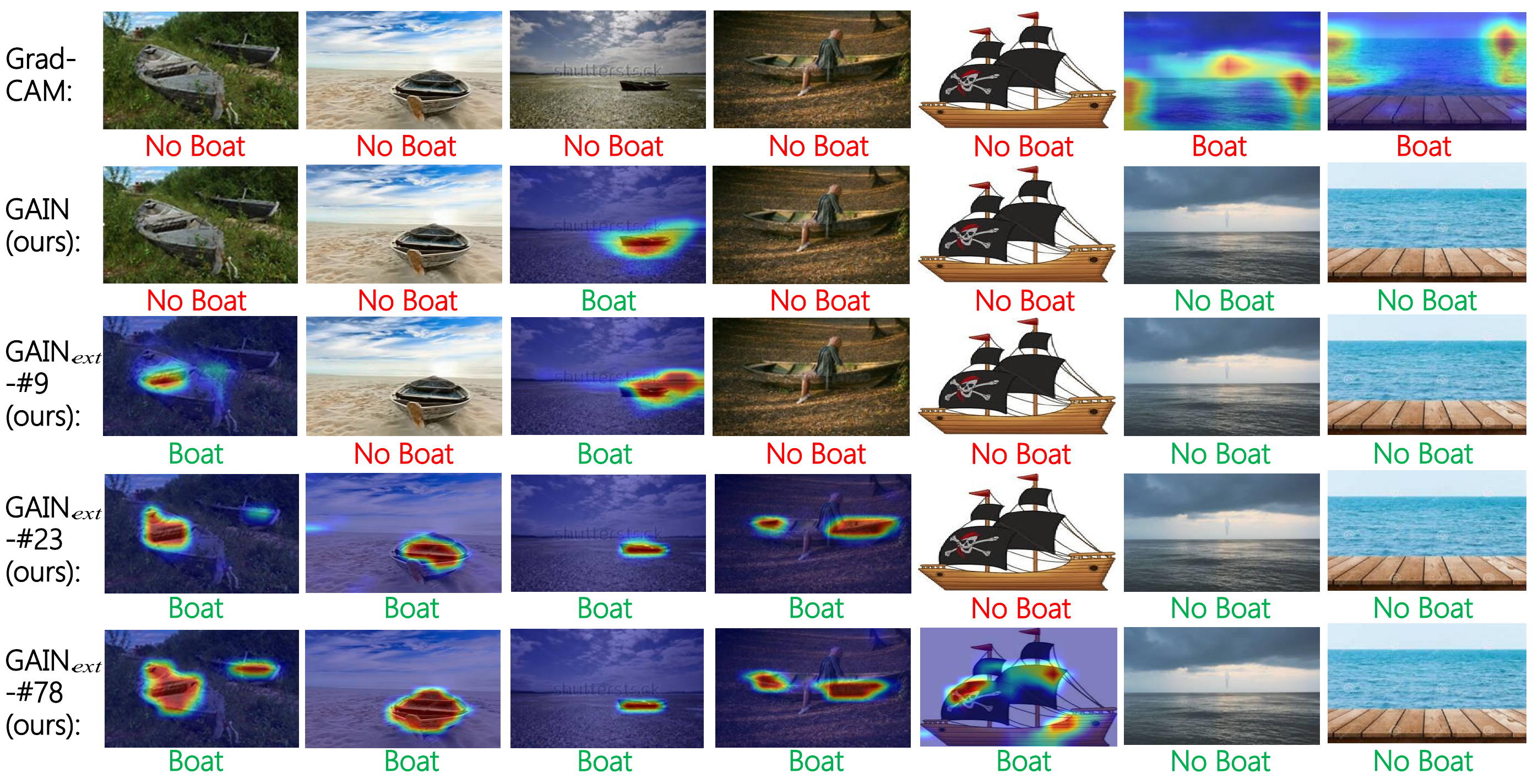} 
\caption{Qualitative results generated by Grad-CAM \cite{grad-cam}, our GAIN and GAIN$_{ext}$ on our \emph{biased boat} dataset. All the methods are trained on Pascal VOC 2012 dataset. \textbf{-\# } denotes the number of pixel-level labels of \emph{boat} used in the training which were randomly chosen from VOC 2012. Attention map corresponding to \emph{boat} shown only when the prediction is positive (i.e. test image contains \emph{boat}).}
\label{fig:Q_result_bias_boat} 
\end{figure*} 

In this section, we design two experiments to verify that our methods have potentials to make the classification network robust to dataset bias and improve its generalization ability by providing guidance on its attention.

\textbf{Boat experiment.} As shown in the Figure \ref{fig:teaser}, the classification network trained on Pascal VOC dataset focuses on sea and water regions instead of boats when predicting there are boats in an image. Therefore, the model failed to learn the right pattern or characteristics to recognize the boats, suffering from the bias in the training set. To verify this, we construct a test dataset, namely \textit{``Biased Boat"} dataset, containing two categories of images: boat images without sea or water; and sea or water images without boats. We collected 50 images from Internet for each scenario. Then we test the model trained without attention guidance, GAIN and GAIN$_{ext}$ described in Section \ref{section:method_human_guided} and \ref{sc:Experiment_comparisons} on this \textit{Biased Boat} test dataset. Results are reported in Table \ref{tb:test_bias_results}. The models are exactly those trained in Sec \ref{sc:Experiment_comparisons}. Some qualitative results are shown in Figure \ref{fig:Q_result_bias_boat}. 

\begin{table}
\begin{center}
\begin{tabular}{cccccc}
\hline
\multirow{2}{*}{Test set} & Grad- & \multirow{2}{*}{GAIN} &\multicolumn{3}{c}{GAIN$_{ext}$ (\# of PL)} \\\cline{4-6}
& CAM &  & 9 & 23 & 78 \\\cline{4-6}\hline\hline
VOC val. & 83\% & 90\% & 93\% & 93\% & 94\%\\ \hline
Boat without water & 42\% & 48\% & 64\% & 74\% & 84\%\\ 
Water without boat & 30\% & 62\% & 68\% & 76\% & 84\%\\ 
Overall & 36\% & 55\% & 66\% & 75\% & 84\%\\ 
\hline
\end{tabular}
\end{center}
\caption{Results comparison of Grad-CAM \cite{grad-cam} with our GAIN and GAIN$_{ext}$ tested on our \emph{biased boat} dataset for classification accuracy. All the methods are trained on Pascal VOC 2012 dataset. \textbf{PL labels} denotes pixel-level labels of \emph{boat} used in the training which are randomly chosen. } 
\label{tb:test_bias_results}
\end{table}

It can be seen that with Grad-CAM \cite{grad-cam} training on VOC 2012, the network has trouble predicting whether there is boat in the image in both of the two scenarios with 36\% overall accuracy. In particular, it generates positive prediction incorrectly on images with only water 70\% of the time, indicating that ``water" is considered as one of the most prominent feature characterizing ``boat" by the network. Using GAIN with only image-level supervision, the overall accuracy on our \emph{boat} dataset has been improved to 55\%, with significant improvement (32\% higher in accuracy, error rate reduced by almost 50\% relatively) on the scenario of ``water without boat''. This could be attributed to that GAIN is able to teach the learner to capture all relevant parts of the target object, in this case, both the boat itself and the water surrounding it in the image. Hence when there is no boat but water in the image, the network is more likely to generate a negative prediction. However with the help of self-guidance, GAIN is still unable to fully decouple boat from water due to the biased training data, i.e. the learner is unable to move its attention away from the water. That is the reason why only 6\% improvement on accuracy is observed in the scenario of ``boat without water''. 

On the other hand with GAIN$_{ext}$ training with small amount of pixel-level labels, similar levels of improvements are observed in both of the two scenarios. With only 9 pixel-level labels for ``boat", GAIN$_{ext}$ obtained an overall accuracy of 66\% on our \emph{boat} dataset, an 11\% improvement compared to GAIN with only self-guidance. In particular significant improvement is observed in the scenario of boats without water.  With 78 pixel-level labels for ``boat'' used in training, GAIN$_{ext}$ is able to obtain 84\% of accuracy on our ``boat'' dataset and performance on both of the two scenarios converged. The reasons behind these results could be that pixel-level labels are able to precisely tell the learner what are the relevant features, components or parts of the target objects hence the actual boats in the image can be decoupled from the water. This again supports that by directly providing extra guidance on attention maps, the negative impact from the bias in training data can be greatly alleviated.

\begin{figure}
\centering
\includegraphics[width=1\linewidth]{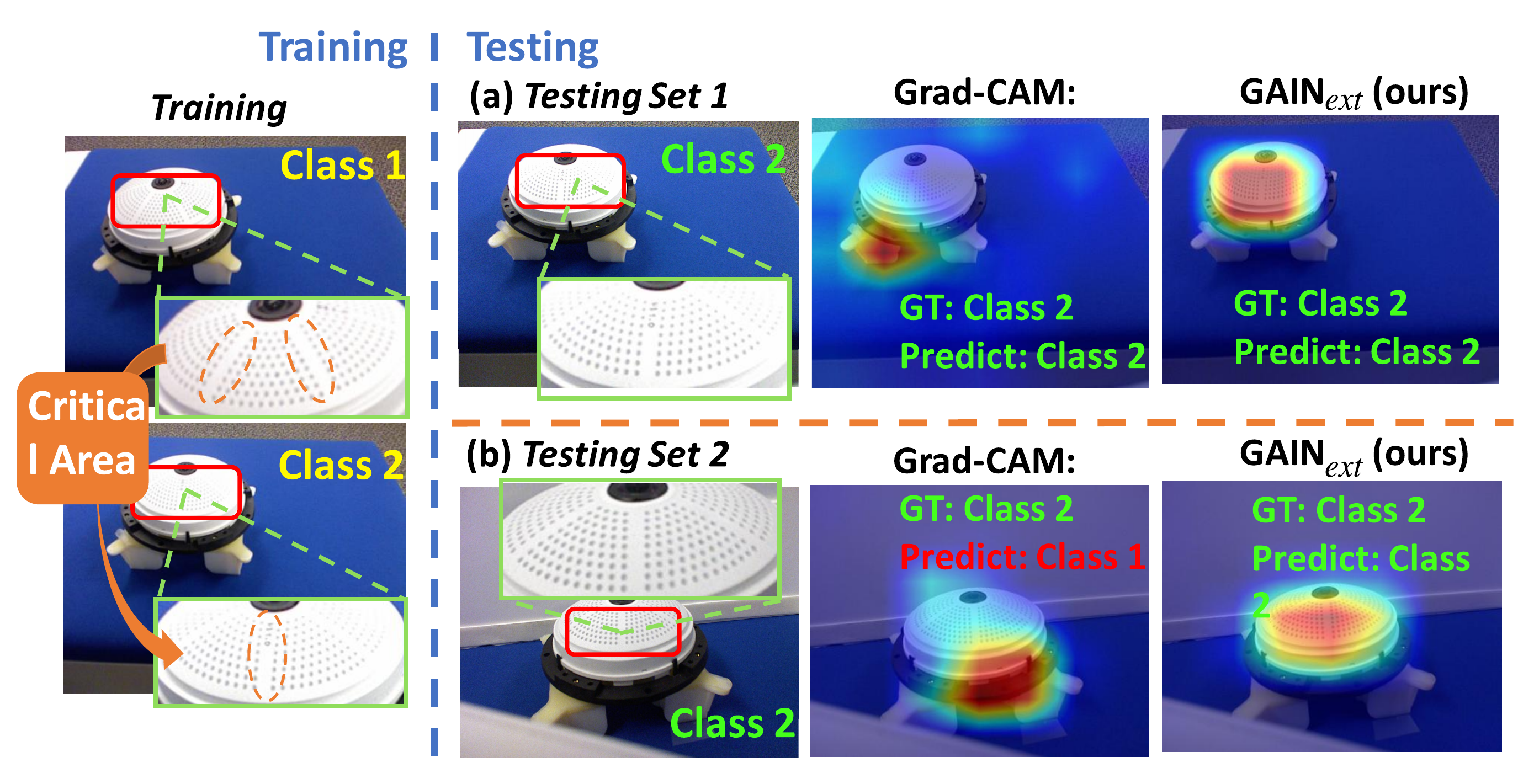} 
\caption{Datasets and qualitative results of our toy experiments. The critical areas are marked with red bounding boxes in each image. \textbf{GT} means ground truth orientation class label.}
\label{fig:toy_experients} 
\end{figure}

\textbf{Industrial camera experiment.} This one is designed for a challenging case to verify the model's generalization ability. We define two orientation categories for the industrial camera which is highly symmetric in shape. As shown in Figure \ref{fig:toy_experients}, only features like gaps and small markers on the surface of the camera can be used to effectively distinguish their orientations. We then construct one training set and two test sets. \textit{Training Set} and \textit{Testing Set 1} are sampled from $\textbf{D}_t$ without overlap. \textit{Testing Set 2} is acquired with different camera viewpoints and backgrounds. There are 350 images for each orientation category in the \emph{Training Set}  resulting in 700 images in total and 100 images each in \emph{Testing Set 1} and \emph{Testing Set 2}. We train VGG-based Grad-CAM and our GAIN$_{ext}$ method on \emph{Training Set}. In training GAIN$_{ext}$, manually drawn bounding boxes (20 for each classes taking up only 5\% of the whole training data) on \textit{critical areas} are used as external supervision. 

At testing stage, though Grad-CAM can correctly classify (very close to 100\% accuracy) the images in \emph{Testing Set 1} where the camera viewpoint and background are very similar to the training set, it only gets random guess results (close to 50\% accuracy) on \emph{Testing Set 2} where images are taken from different shooting camera viewpoint with different background. This is due to the fact that there is severe bias in the training dataset and the learner fails to capture the right features (\textit{critical area}) to separate the two classes. On the contrary, using GAIN$_{ext}$ with small amount of images with bounding-box labels (5\% of the
whole training data), the network is able to focus its attention on the area specified by the bounding box labels hence better generalization can be observed when testing with \emph{Testing Set 2}. Although shooting camera viewpoint and scene background are quite different from the training set, the learner can still correctly identify the critical area on the camera in the image as shown in last column second row in \ref{fig:toy_experients}, and hence correctly classified all images in both \emph{Testing Set 1} and \emph{Testing Set 2}. The results again suggest that our proposed GAIN$_{ext}$ has the potential of alleviating the impact of biases in training data, and guiding the learner to generalize better.

\section{Conclusions}

We propose a framework that provides direct guidance on the attention map generated by a weakly supervised learning deep neural network in order to teach the network to generate more accurate and complete attention maps. We achieve this by making the attention map not an afterthought, but a first-class citizen during training. Extensive experiments demonstrate that the resulting system confidently outperforms the state of the art without the need for recursive processing during run time. The proposed framework can be used to improve the robustness and generalization performance of networks during training with biased data, as well as the completeness of the attention map for better object localization and segmentation priors. In the future it may be illuminating to deploy our method on other high-level tasks than categorization and to explore for instance how a regression-type task may benefit from better attention.

\section{Acknowledgments}
This paper is based primarily on the work done during Kunpeng Li' s internship at Siemens Corporate Technology. This research is supported in part by the NSF IIS award 1651902, ONR Young Investigator Award N00014-14-1-0484 and U.S. Army Research Office Award W911NF-17-1-0367.

{\small
\bibliographystyle{ieee}
\bibliography{egbib}
}

\end{document}